\documentclass{article}
\pdfoutput=1
\usepackage{arxiv}
\usepackage{lineno,hyperref}
\usepackage{amsmath}
\usepackage{graphicx}
\usepackage{amsfonts}
\usepackage{longtable}
\usepackage{adjustbox}
\usepackage{hhline}
\usepackage{multirow}
\usepackage{algorithm}
\usepackage{algorithmic}
\usepackage{slashbox}
\usepackage[T1]{fontenc}
\usepackage{lscape}
\usepackage{xcolor}
\usepackage{appendix}

\usepackage[utf8]{inputenc} 
\usepackage[T1]{fontenc}    
\usepackage{url}            
\usepackage{booktabs}       
\usepackage{amsfonts}       
\usepackage{nicefrac}       
\usepackage{microtype}      
\usepackage{lipsum}

\title{Diverse and Styled Image Captioning Using SVD-Based Mixture of Recurrent Experts}

\author{Marzieh Heidari \\
  Department of Mathematics and Computer Science \\
  Amirkabir University of Technology (Tehran Polytechnic)\\
  Iran\\
   \And
  Mehdi Ghatee\footnote{The corresponding author} \\
  Department of Mathematics and Computer Science \\
  Amirkabir University of Technology (Tehran Polytechnic)\\
  Iran\\
  \texttt{ghatee@aut.ac.ir}\\
     \And
   Ahmad Nickabadi \\
  Department of Computer Engineering \\
  Amirkabir University of Technology (Tehran Polytechnic)\\
  Iran\\
     \And
   Arash Pourhasan Nezhad \\
 Department of Mathematics and Computer Science \\
  Amirkabir University of Technology (Tehran Polytechnic)\\
  Iran\\
}

\begin{document}
\maketitle
		\begin{abstract}
			With great advances in vision and natural language processing, the generation of image captions becomes a need. In a recent paper, Mathews, Xie and He \cite{RN28}, extended a new model to generate styled captions by separating semantics and style. In continuation of this work, here a new captioning model is developed including an image encoder to extract the features, a mixture of recurrent networks to embed the set of extracted features to a set of words, and a sentence generator that combines the obtained words as a stylized sentence. The resulted system that entitled as Mixture of Recurrent Experts (MoRE), uses a new training algorithm that derives singular value decomposition (SVD) from weighting matrices of Recurrent Neural Networks (RNNs) to increase the diversity of captions. Each decomposition step depends on a distinctive factor based on the number of RNNs in MoRE. Since the used sentence generator gives a stylized language corpus without paired images, our captioning model can do the same. Besides, the styled and diverse captions are extracted without training on a densely labeled or styled dataset. To validate this captioning model, we use Microsoft COCO which is a standard factual image caption dataset. We show that the proposed captioning model can generate a diverse and stylized image captions without the necessity of extra-labeling. The results also show better descriptions in terms of content accuracy.
		\end{abstract}
		
	\section{Introduction}
	Generating human-like captions for images automatically, namely, image captioning, has risen as an interdisciplinary research issue at the crossing point of computer vision and natural language processing \cite{RN28,yang2019visual,yan2020image,xu2019scene, RN33, RN54, RN34}. It has numerous imperative industrial applications, such as assistant facilities for visually impaired individuals, visual knowledge in chatting robots, and photo sharing on social media. For producing genuine human-like image captions, an image captioning framework is required to understand the visual content of input image and write captions with proper linguistic properties. Nonetheless, most existing image captioning frameworks center around the vision side that describes the visual content in an objective and neutral manner (factual captions), while the language side, e.g. linguistic style, is regularly disregarded. In fact, linguistic style \cite{RN55} is an essential factor that reflects human personality \cite{RN56}, influences purchasing decisions\cite{RN57} and fosters social interactions \cite{RN58}. The different styles in image captioning is also an important problem that has been referred by \cite{RN58, RN59, RN33}.
	
	Generating styled, diverse and accurate captions for an image is an open challenge. For gaining diversity in generated captions some works require manually created, densely labeled, image caption datasets\cite{RN29,yatskar2014see,yang2017dense}, some use GANs\cite{RN42} to achieve diversity which mostly suffers from poor accuracy. Also For gaining style most works use styled datasets\cite{RN22, RN44}.
	
	We address the problem of necessity of gathering styled and densely labeled datasets to generate styled and diverse captions by presenting a novel unified model architecture that can generate styled, diverse and accurate captions without using extra labels and trained only on a standard image captioning dataset and a styled corpus.
	
	Central to our approach is reducing the requirement of immense, densely labeled and styled dataset for image captioning. We propose a model for generating styled, diverse, and semantically relevant image captions containing an image embedder, a Term Generator, and a Sentence Generator. Image embedder is the one before the last layer of a pre-trained CNN that takes an image as input and outputs the visual features of the image. Term Generator is an MoRE that is responsible for diversity. Each RNN expert generates a specific word sequence. During the training of each RNN in Term Generator, at the end of each epoch, we filter out a part of deep network weights using SVD decomposition to generate diverse captions without an extra label. Previously SVD has been used for network compression \cite{goetschalckx2018efficiently} and overfitting controlling \cite{bejani2020NN}, but this is the first time SVD is used for diverse captioning. Sentence Generator is responsible for controlling style. It learns style from a corpus of stylized text without aligned images.
	We evaluate our model on COCO dataset\cite{RN49}. After the evaluation of sentences by each Term Generator expert, we extracted the vocabulary from their sentences. The vocabulary sets are different in both of the lengths and their content.
	
	Our contribution is developing a new model that can generate styled, diverse, and accurate captions for images without training on a densely labeled dataset.
	
	To discuss on some related works, in earlier image captioning studies, template-based models \cite{kulkarni2013babytalk,elliott2015describing} or retrieval-based models \cite{devlin2015language} were commonly used. The template-based models distinguish visual concepts from a given image and fill them into some well-defined formats to make sentences. In this way, the generations suffer from lack of diversity. The retrieval-based models discover the foremost reliable sentences from existing ones and are not able to produce new descriptions.
	
	On the other hand, end-to-end trainable image captioning models are the result of recent advances in deep learning and the release of large scale datasets such as COCO \cite{RN15} and Flickr30k\cite{RN32}. Most modern image captioning models use the encoder-decoder framework \cite{yang2019visual, RN33, RN34, RN35, RN36}, where a convolutional neural network (CNN) encodes the input image in vector space feature embeddings which are fed into an RNN. The RNN takes the image encoding as input and the word generated in the current time step to generate a complete sentence one word at a time. Maximum likelihood estimation is typically used for training. It has been shown that attention mechanisms\cite{RN33, RN26, RN25} and high-level attributes/concepts \cite{RN37, RN38} can help image captioning. Recently, reinforcement learning has been introduced for image captioning models to directly optimize task-specific metrics \cite{RN39, RN40}. Some works adopt Generative Adversarial Networks (GANs)\cite{RN42}  to diverse or generate human-like captions \cite{RN41}. Some works have been also adapted by weakly-supervised training methods \cite{zheng2018weakly} for making richer captions.
	\subsection{Diverse Image Captioning } \label{diverseCaptions}
	Generating diverse captions for images and videos has been studied in the recent years\cite{dai2017towards,jain2017creativity,shetty2017speaking,vijayakumar2016diverse,wang2017diverse,zhang2018more}.  Some techniques such as GAN-based methods \cite{shetty2017speaking,dai2017towards} and VAE-based methods \cite{jain2017creativity,wang2017diverse} are used to improve the diversity and accuracy of descriptions. Some others \cite{chatterjee2018diverse,dai2017towards} studied generating descriptive paragraphs for images. Also in \cite{deshpande2018diverse}, a method has been proposed to apply the part-of-speech of words in generated sentences. In \cite{mao2018show}, generated sentences can contain words of different topics. The research of \cite{RN29}, generates descriptions for each semantic informative region in images. In addition, in \cite{zheng2019intention}, a particular guiding object that is presented in the image is chosen to be necessarily presented in the generated description. Most of these approaches require additional labels in a dataset, e.g. Visual Genome \cite{krishna2017visual}. In what follows, we propose a new scheme without the necessity to give additional labels.
	\subsection{Stylized Image Captioning }
	Stylized image captioning aims at generating captions that are successfully stylized and describe the image content accurately. The works proposed for tackling this task
	can be divided into two categories: models using parallel stylized image-caption data (the supervised mode) \cite{RN22,RN43,RN23,RN21} and models using non-parallel stylized corpus (the semi-supervised mode) \cite{RN44,RN28}. SentiCap \cite{RN22} handles the positive/negative styles and proposes to model word changes with two parallel Long Short Term Memory (LSTM) networks and word-level supervision. StyleNet \cite{RN44} handles the humorous/romantic styles by factoring the input weight matrices to contain a style specific factor matrix. SF-LSTM \cite{RN43} experiments on the above four caption styles and proposes to learn two groups of matrices to capture the factual and stylized knowledge, respectively. You et al. \cite{RN21} proposed two simple methods to inject sentiments into image captions. They could control the sentiment by providing different sentiment labels. See and Chat \cite{chen2019see} in the first step retrieved the visually similar images form their dataset which contains 426K images with 11 million associated comments, to input image using k nearest neighbors and then ranked their comments to get the most relevant comment for the input image. There are more than 25 comments for each image on average. Since the comments of the dataset are styled, the resulted caption was styled too.
	
	\section{A New Model with a Mixture of Experts}\label{The Proposed Model}
	As stated in Section~\ref{diverseCaptions}, the current image captioning methods need extra labels on the training data to generate stylized captions. In this section, we propose a novel image captioning model that applies diversity and style to the generated captions without requiring additional labels. The goal of our proposed model is to take an image, a target style, and a diversity factor as input and generate a sentence within the target style. To find better results, we use an ensemble of neural networks. To this aim, different kinds of ensemble models can be used. For example, ensemble of neural networks in tree structure \cite{abpeykar2019NCAA,abpeykar2019ensemble} or mixture of experts \cite{abbasi2016regularized,pashaei2019convolution}  can be followed. We follow a Mixture of Recurrent Experts (MoRE) in what follows.
	
	The architecture of the proposed model is illustrated in Figure~\ref{fig2}. The model is comprised of three basic components, i.e., an image encoder, a Term Generator, and a Sentence Generator. The image encoder is a CNN that extracts visual features from the input image. The Term Generator is a mixture of some RNNs that takes visual features extracted by CNN and SVD factors as input and gives a sequence of semantic terms. The Sentence Generator is an attention-based RNN that takes this sequence of semantic terms and the target style and decodes them into a sentence in natural language with a specific style that describes the image. Each RNN of the Term Generator component has a specific SVD factor. The SVD factor shows what portion of the RNN weight matrix is saved during training. At test time, the SVD factors determine which one of the experts is responsible to generate the sequence of semantic terms from visual features.
	
	\begin{figure}[tbp]
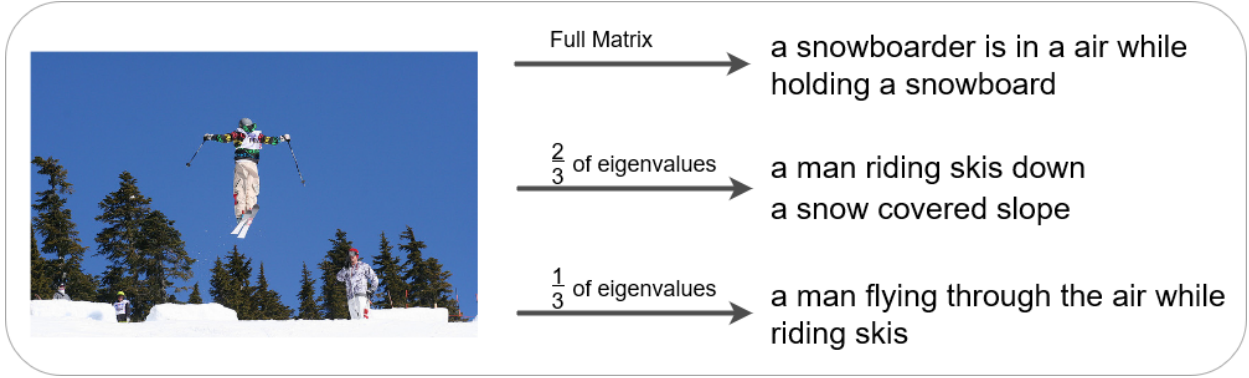

		\centerline{\includegraphics[width=1\textwidth]1}
		\caption{The results of applying different SVD factors to approximate weight matrix. The image is from COCO \cite{RN49}. }
		\label{fig1}
	\end{figure}
	\begin{figure*}[htbp]
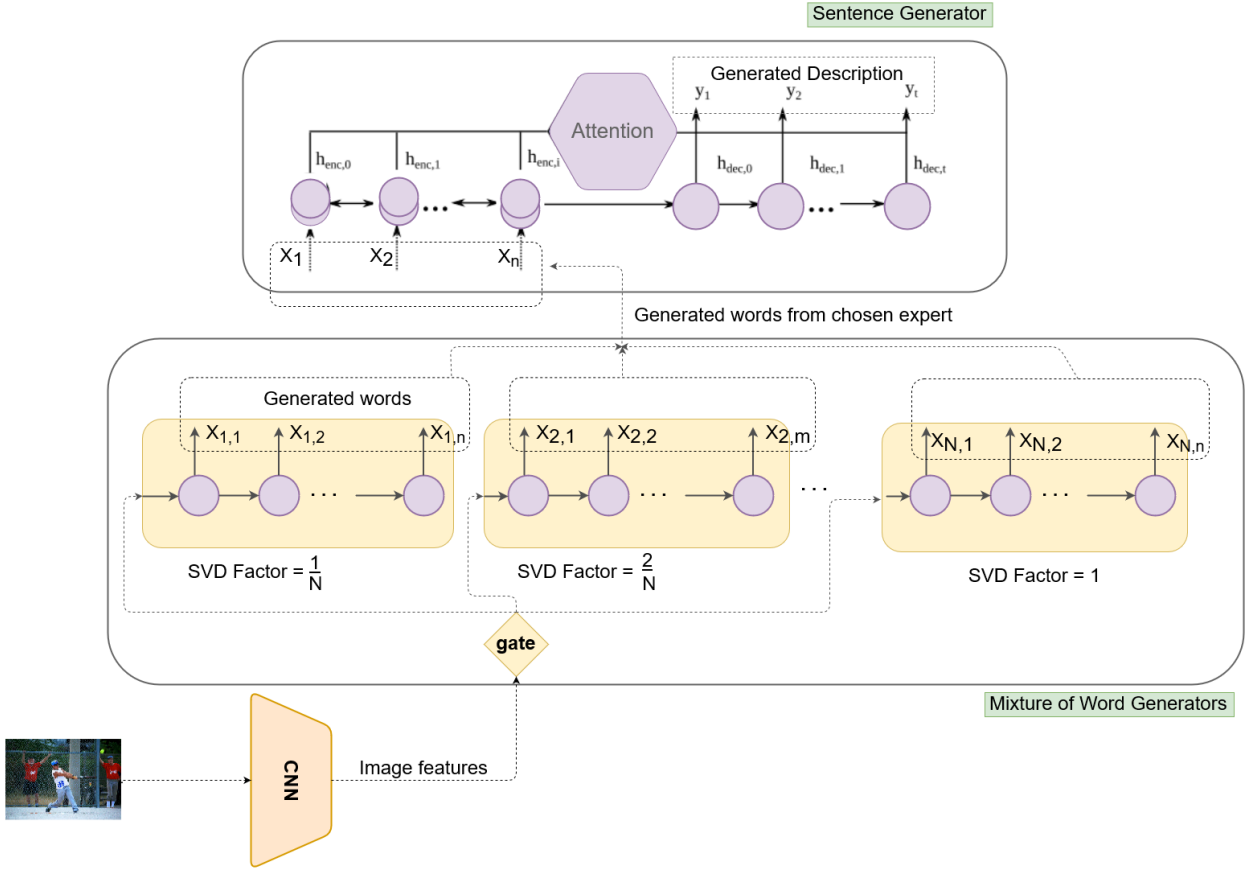

		\centerline{\includegraphics[width=1\textwidth]2}
		\caption{The architecture of MoRE to generate the multi-style captions with target style. SVD Factor indicates what portion of matrix rank is remained in the deep network. The output of each expert is supposed as the input of Sentence Generator.The attention is from SemStyle \cite{RN28}}
		\label{fig2}
		
	\end{figure*}
	Each Term Generator uses a different SVD factor to cause diversity in extracted words and consequently diversity in the generated captions. Moreover, we designed a two-stage learning strategy to train the Term Generator networks and the Sentence Generator network separately. We train the Term Generators on a dataset of image caption pairs and the Sentence Generator on a corpus of styled text data such as romantic novels. 
	\subsection{Image Encoder}
	This module encodes any image $I$ to get features utilizing a deep CNN. Previous studies use different types of image features. The image features could be local visual features for every semantic segment of image \cite{RN34} or a static global representation of the image \cite{RN33}. A visual context vector is obtained by directly using the static feature or calculating adaptively with a soft-attention mechanism\cite{RN34,he2019vd}. In this paper, we use the static features to remain consistency with the previous works. The image features are extracted from the second last layer of the Inception-v3\cite{RN48} of CNN pre-trained on ImageNet \cite{russakovsky2015imagenet}.
	\subsection{Term Generator}
	The Term Generator network is an MoRE that maps an input image, denoted by $I$ to an ordered sequences of semantic terms $x = {x_{1}, x_{2}, ..., x_{M}}, x_{i} \in V^{word} $. Each semantic term is a word with a part-of-speech tag. These words indicate the objects, scene, and activity in the image. This generator should completely capture the visual semantics and should be independent of linguistic style; because Sentence Generator is responsible for applying a style to caption.
	
	Our MoRE is trained by an SVD based approximation method	inspired by\cite{xue2013restructuring}.
	For a learnt weight matrix $ W $, by approximating by an SVD, one can find
	\begin{align}
	W_{m\times n} =U_{m\times n} \Sigma_{n\times n} V^{T}_{n\times n}  \label{f2}
	\end{align}
	where $\Sigma$ is a diagonal matrix with singular values on the diagonal in the decreasing order. $m$ columns of $U$ and $n$ columns of $V$ are called the left-singular vectors and the right-singular vectors of $A$, respectively. By approximating $A$ by the greatest $k$ components of this decomposition, one can substitute the following instead of $A:$      
	\begin{align}
	W_{m\times n} =U_{m\times l} \Sigma_{l\times l} V^{T}_{l\times n}  \label{f3}
	\end{align}              
	By changing $l$, the different variations of the weighting matrix can be defined for Term Generator networks and each approximation interprets any image differently. Thus, the outputs will be variant. In our model, for $i^{th}$ RNN, we define a diversity factor $k=\frac{i}{R}$ where $R$ is the number of RNNs in MoRE. For each expert of MoRE, $l=[k*rank(W)]$ shows the portion of the principal components of matrix $W$ that remains in the learning model. Noe that $rank(W)$ denotes the rank of $W.$	The effect of different diversity factors on the generating the sequence of semantic terms is shown in Fig~\ref{fig1}.	
	
	The architecture of all experts of MoRE are similar and is a CNN+RNN inspired by Show and Tell \cite{RN33}, see the middle part of Fig.~ \ref{fig2}.  The image feature vector passes through a densely connected layer and then through an RNN with Gated Recurrent Unit (GRU) cells\cite{cho2014learning}. The word list $x$ is shorter than a full sentence, which speeds up training and alleviates the effect of forgetting long sequences. At each time-step $t$, there are two inputs to the GRU cell. The first is the previous hidden state summarizing the image $I$ and word history $x_{1}, ..., x_{t-1}$. The second is the GloVe embedding vector $E_{x_{t}}$ of the current word. A fully connected layer with softmax takes the output $h_{t} $ and produces a categorical distribution for the next word in the sequence $x_{t+1}$. Argmax decoding can be used to recover the entire word sequence from the conditional probabilities. See Eq. (1) in \cite{RN28}. We set $ x_{1} $ as the beginning-of-sequence token and terminate when the sequence exceeds a maximum length or when the end-of-sequence token is generated.

	\subsection {Sentence Generator}
	The Sentence Generator, shown in the upper part of
	Fig.~\ref{fig2}, maps the sequence of semantic terms to a sentence with a specific style. For example, given the word list “girl”, “posture”, “refrigerator”, and “DESCRIPTIVE” as the requested style, a suitable caption is “A girl standing in a kitchen beside a refrigerator.” Also the same list of words with “STORY” as the expected style as the input is “I saw the girl standing in the kitchen, and I was staring at the refrigerator”. Given the list of words $ x $ and a target style $z,$ we generate an output caption $y = {y_{1}, y_{2},..., y_{t}, ..., y_{L}},$ where $y_{t} \in V^{out} $ and $ V^{out} $ is the output word vocabulary. To do so, the idea of \cite{RN28} is used by considering an RNN sequence-to-sequence sentence generator network with attention over the input sequence. This is an auto-encoder that maps input word sequence to a vector space and decodes the sequence to a sentence to describe the image in a suitable style. Encoder component for sequence $ x $ consists of a GloVe vector embedding followed by a batch normalization layer \cite{RN52} and a bidirectional RNN \cite{schuster1997bidirectional} with GRU cells. The Bidirectional RNN \cite{RN28} is implemented as two independent RNNs. They run in opposite directions with shared embeddings. For details, we refer to Eq. (4) of \cite{RN28}.
	
	\begin{figure*}[htbp]
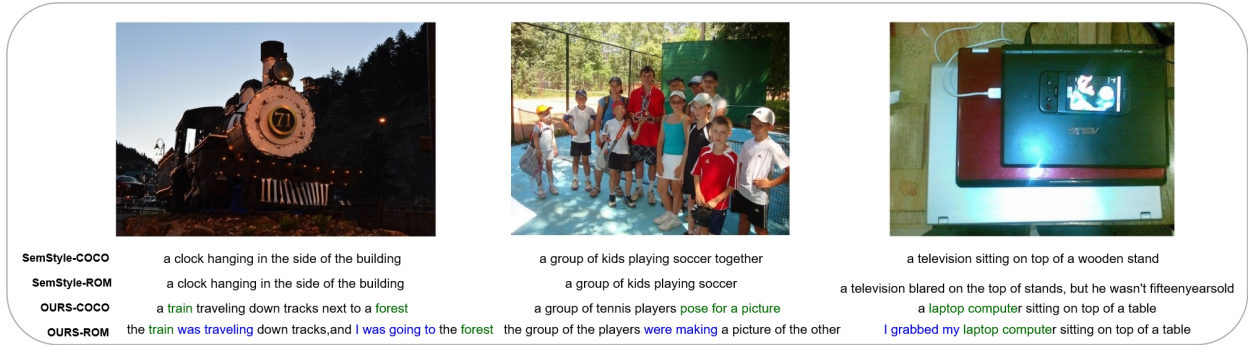

		\centerline{\includegraphics[width=1\textwidth]0}
		\caption{Some results to compare the description of our model with that of SemStyle\cite{RN28}. Green color shows enhancement in accuracy and blue color indicates the improvement of style. The images are from COCO\cite{RN49}. }
		\label{fig3}
	\end{figure*}
	\section{Experimental Setup\label{experiment}}
	
	We conduct experiments on publicly available image caption dataset, Microsoft COCO \cite{RN15}. COCO is a large image captioning dataset, containing 82783, 40504, and 40775 images for training, validation, and test, respectively. Each image is labeled with 5 human-generated descriptions for image captioning. All labels are converted to lower case and tokenized. We use Semstyle training and testing split sets \cite{RN28} for both factual and stylized captions. 
	
	We consider 9 baseline methods and compare them with 5 variants of our proposed captioning model. The considered baselines are as the following:
	\begin{itemize} 
		\item Show and Tell \cite{RN33} constructed a CNN as an encoder and an RNN as the decoder. 
		\item Neural Talk \cite{karpathy2015deep} used the images and their regions alignments to captions to learn and to generate descriptions of image regions.  
		\item StyleNet \cite{RN44} originally has been trained on FlickrStyle10K \cite{RN44}. The implementation of StyleNet and  StyleNet-COCO are from \cite{RN44}. This reimplementation makes the trained datasets match and consequently makes the approaches comparable. StyleNet generates styled captions, while StyleNet-COCO generates descriptive captions. 
		\item Neural-storyteller \cite{RN45} is a model trained on romance text (from the same source as ours).
		\item JointEmbedding maps images and sentences to a continuous multi-modal vector space \cite{RN46} and uses a separate decoder, that has been trained on the romance text, to decode from this space.
		\item SemStyle \cite{RN28} is our reference to develop the model and maps image features to a word sequence and then maps the sequence to a caption.
		\item SGC \cite{xu2019scene} applies the Scene Graph Captioner (SGC)
		framework for the image captioning task.
		\item Hierarchical Attention \cite{yan2020image} uses a hierarchical attention model by
		utilizing both of the global CNN features and the local object features for more effective feature representation and reasoning in image captioning. 
		\item VSV-VRA-POS \cite{yang2019visual} the adapts the language models for
		word generation to the specific syntactic structure of sentences and visual skeleton of the image.
	\end{itemize}
	The variants of our proposed captioning model are as the following:
	\begin{enumerate} 
		\item \textbf{Shuffled words} model is genuinely base-line that during the training of the Sentence Generator, the input words are out of order. This gives a little noise to input and the results are less overfitted.
		\item \textbf{Shuffles words+batch normalization} model is "Shuffled words" model that uses a batch normalization layer after the embedding layer of Sentence Generator. This makes features more general and consequently more general captions.
		\item \textbf{Shuffles words+Glove+batch normalization} model uses freezed weights of Glove pre-trained embedding.
		\item \textbf{Full model} applies a specific drop-out to embedding layers. 
		\item \textbf{Kaldi GRU} is a full model that uses Kaldi Speech Recognition \cite{povey2011kaldi} GRUs as encoder and decoder in Sentence Generator instead of typical GRUs.
	\end{enumerate}
	\subsection{Evaluation Metrics}
	We use two types of metrics to evaluate the proposed image captioning model. The first type is automatic relevance metrics. In this part, similar to \cite{RN28}, we consider captioning metrics including BLEU \cite{RN16}, METEOR\cite{RN17}, ROUGE\_L \cite{lin2004rouge}, and CIDEr \cite{RN14} which are based on n-gram overlap and SPICE \cite{RN19} based on f-score over semantic tuples extracted from COCO reference sentences \cite{RN49}. As the second type of metric, we consider automatic style metrics. In this part, we measure how often a generated caption has the correct target-style according to a pre-trained style classifier. The CLassifier Fraction (CLF) metric \cite{RN28}, is the fraction of generated captions classified as styled by a binary classifier. This classifier is logistic regression with 1,2-gram occurrence features trained on styled sentences and COCO training captions. Its cross-validation precision is 0.992 at a recall of 0.991. 

	\subsection{Training details}
	In our experiments, the model is optimized with Adam \cite{RN47}. The learning rate is set to 1e-3.  We clip gradients to [-5, 5] and apply dropout to image and sentence embeddings. The mini-batch size is 64 for both the Term Generator and the Sentence Generator.  Both the Term Generator and Sentence Generator use separate 512-dimensional GRUs and word embedding vectors. The Term Generator has a vocabulary of 10000 words while the Sentence Generator has two vocabularies: one for encoder input another for the decoder – both vocabularies have 20000 entries to account for a broader scope. The number of intersecting words between the Term Generator and the Sentence Generator is 8266 with both datasets, and 6736 without. Image embeddings come from the second last layer of the Inception-v3 CNN\cite{RN48}  and are 2048 dimensional.
	
	We used Glove \cite{RN50} frozen weights for embedding layers of both Term Generator and Sentence Generator of Semstyle's baseline. The model suffered from overfitting so we adopt the following regularization techniques in order to fix this problem: \begin{itemize}
		\item  Instead of normal drop-out, we used embedding-specific drop-out proposed by Merity et al. \cite{RN51}.
		\item We adopt batch normalization \cite{RN52} for both modules of the model including Term Generator and Sentence Generator.
		\item For Term Generator, we used weight decay \cite{RN53} with a coefficient of 1e-6.
		\item In training Sentence Generator instead of feeding ordered semantic terms, we shuffled them so the model learns to generate sentences from unordered semantic terms that improve the generalization.
	\end{itemize}
	We trained three Term Generator experts. Each expert of MoRE applies SVD on the weighting matrices in the Term Generator model, and afterward reconstruct the model based on the inherent sparseness of the original matrices. For each expert, we saved $[k*rank(W)]$ of principal components of learned weight matrix $W$ for SVD approximation. This approach has been used by \cite{xue2013restructuring} for the neural network training process. After reconstruction, the accuracy decreases but the final classification results improve. After every epoch, the reduced weights are replaced in the model. Afterward, we fine-tune the reconstructed model using the back-propagation method to receive better accuracy.
	\subsection{Results}
	The results of measurements are presented in Table \ref{table1}. Table \ref{table2} shows the results of the automatic metrics on caption style learned from romance novels. This comparison is similar to \cite{RN28}. Also, our full model generates descriptive captions. It accomplishes semantic relevance scores comparable to the Show and Tell \cite{RN33}, with a SPICE of 0.166 vs 0.154, and BLEU-4 of 0.252 vs 0.238. Thus utilizing semantic words is a competitive way to distill image semantics. Really, the Term Generator and the Sentence Generator constitute a compelling vision-to-language pipeline. Additionally, our model can create different caption styles, where the CLF metric for captions classification is 99.995\%, when the target style is descriptive. Figure~\ref{fig3} demonstrates some quantitative results for different styles alongside results of the same images generated by SemStyle which is the most similar approach to ours.
	
	In addition, our model generates styled captions in 74.1\% of cases, based on CLF. SPICE score is 0.145 which is better than the presented baselines. Results are shown in Table~ \ref{table2}.
	
    As one can see in Table \ref{table2} there are three models (SGC \cite{xu2019scene}, VSV-VRA-POS \cite{yang2019visual} and Hierarchical Attention \cite{yan2020image} ) with better relevance scores compared with our work. These three works are presented to show the impact of using unpaired captions and a unified model for multi-style captioning. The trade-off for such a model would be the loss of relevance scores. Because when you have only one objective (which is generating similar captions to ground-truth) the similarity score would be higher compared with the works that peruse more objects. Since, our model tries to improve the captions for all styles, our method can outperform other one-objective methods, in real situations.
	
	\subsection{Component Analysis}
	According to Table~\ref{table3} and Table~\ref{table4} the poorest result is gotten when Kaldi GRU is used instead of regular GRU cells. By shuffling the input words of Sentence Generator input in training, we can see a little improvement which is the result of decreasing overfit caused by adding this noise. Adding a Batch Normalization layer boosts the results so that in some metrics such as BLEU3, BLEU4, and Cider the best result is achieved by this model. Adding Glove frozen weights only improves styled captioning by increasing the number of styled captions by 10\%. But it decreases other scores slightly. This score-dropping is the result of overfitting so in the full model, we added embedding-specific dropout layer to fix this issue. As a result, relevance scores boost up again and in addition, another 12\% added to the styled caption on style evaluation.                
	\begin{figure}[tbp]
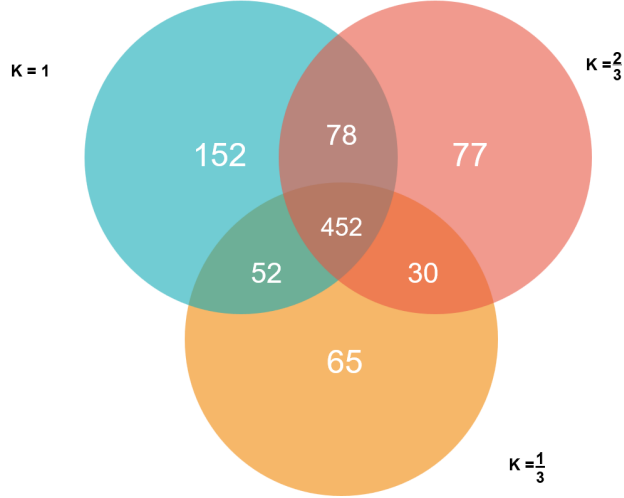

		\centerline{\includegraphics[width=0.5\textwidth]3}
		\caption{ Words scattering for generated captions by three models with three different SVD factors on the weighting matrices. The numbers are the count of words and different colors on different Term Generators inside the MoRE. }
		\label{fig4}
	\end{figure}
	\scriptsize
	\begin{table*}[]
		\centering    \caption{Evaluating caption relevance on the COCO dataset.}
		\label{table1}
		\scriptsize
		\begin{tabular}{p{1.3cm}|p{0.8cm}|p{0.7cm}|p{0.7cm}p{0.7cm}p{0.7cm}p{0.7cm}p{0.7cm}p{0.7cm}p{0.7cm}p{0.7cm}}
			\hline
			model        & \rotatebox[origin=c]{270}{Multi-Style}        & \rotatebox[origin=c]{270}{Unpaired}     & \rotatebox[origin=c]{270}{BLEU1}          & \rotatebox[origin=c]{270}{BLEU2}          & \rotatebox[origin=c]{270}{BLEU3}          & \rotatebox[origin=c]{270}{BLEU4}          & \rotatebox[origin=c]{270}{METEOR}         & \rotatebox[origin=c]{270}{ROUGE\_L}       & \rotatebox[origin=c]{270}{CIDEr}& \rotatebox[origin=c]{270}{SPICE}\\ \hline
			Show and Tell \cite{RN33}     & no    & no           & 0.667          &                &                & 0.238          & 0.224          &                & 0.772          & 0.154          \\
			Neural Talk\cite{karpathy2015deep}      & no     & no           & 0.625          & 0.45           & 0.23           & 0.23           & 0.195          &                & 0.66           &                \\
			StyleNet (COCO) \cite{RN44}   & no           & yes          & 0.643          &                &                & 0.212          & 0.218          &                & 0.664          & 0.135          \\
			SGC \cite{xu2019scene} &no& no& 0.679 &0.493&0.347&0.243&0.222&0.754&0.488&-\\
			Hierarchical Attention \cite{yan2020image}& no &no & 0.7261 & 0.5277 & 0.3780&0.2724&0.2473&0.8814&0.5604&-\\
			VSV-VRA-POS \cite{yang2019visual} & no &no&0.782 &0.619 &0.477 &0.368 & 0.277& 0.572& 1.159&0.207\\ 
			SemStyle (COCO) \cite{RN28}     & yes  & yes          & 0.653          & 0.478          & 0.337          & 0.238          & 0.219          & 0.482          & 0.769          & 0.157          \\ \hline
			\textbf{MoRE (ours)} & \textbf{yes} & \textbf{yes}& \textbf{0.679} & \textbf{0.501} & \textbf{0.356} & \textbf{0.252} & \textbf{0.226} & \textbf{0.501} & \textbf{0.844} & \textbf{0.166} \\ \hline
		\end{tabular}
		\normalsize
	\end{table*}
	\begin{table*}[]\scriptsize
		\centering    \caption{Evaluating styled captions with automated metrics.}
		\label{table2}
		\begin{tabular}{c|c|c|cc}
			\hline
			Model              & Multi-Style & Unpaired & SPICE & CLF   \\ \hline
			StyleNet  \cite{RN44}         & no          & yes      & 0.010 & 0.415 \\
			neural-storyteller\cite{RN45} & no          & no       & 0.057 & 0.983 \\
			JointEmbedding \cite{RN46}    & no          & no       & 0.046 & 0.99  \\
			SemStyle (ROM) \cite{RN28}    & yes         & yes      & 0.144 & 0.589 \\ \hline
			MoRE(ours)         & yes         & yes      & 0.145 & 0.741 \\ \hline
		\end{tabular}
	\end{table*}
	\begin{table*}[]
		\centering    \caption{Experimental result for descriptive style.}
		\label{table3}
		\scriptsize
		\begin{tabular}{p{3cm}|p{0.6cm}p{0.6cm}p{0.6cm}p{0.6cm}p{0.6cm}p{0.6cm}p{0.6cm}p{0.6cm}}
			\hline
			model        &  \rotatebox[origin=c]{270}{BLEU1}          & \rotatebox[origin=c]{270}{BLEU2}          & \rotatebox[origin=c]{270}{BLEU3}          & \rotatebox[origin=c]{270}{BLEU4}          & \rotatebox[origin=c]{270}{METEOR}         & \rotatebox[origin=c]{270}{ROUGE\_L}       & \rotatebox[origin=c]{270}{CIDEr}& \rotatebox[origin=c]{270}{SPICE}\\ \hline
			kaldi GRU                                    & 0.606 & 0.437 & 0.297 & 0.201 & 0.204  & 0.464    & 0.632 & 0.145 \\
			shuffles words                               & 0.655 & 0.486 & 0.347 & 0.247 & 0.223  & 0.497    & 0.813 & 0.163 \\
			shuffles words+batch normalization           & 0.664 & 0.494 & 0.357 & 0.258 & 0.226  & 0.5      & 0.837 & 0.164 \\
			shuffles words+embedding+batch normalization & 0.657 & 0.49  & 0.353 & 0.254 & 0.224  & 0.498    & 0.83  & 0.163 \\
			full model                                   & 0.679 & 0.501 & 0.356 & 0.252 & 0.226  & 0.501    & 0.844 & 0.166 \\ \hline
		\end{tabular}
	\end{table*}
	\begin{table*}[]\scriptsize
		\centering    \caption{Experimental result for romance style.}
		\label{table4}
		\begin{tabular}{l|ll}
			\hline
			model                                        & SPICE & CLF   \\ \hline
			shuffles words                               & 0.147 & 0.553 \\
			shuffles words+batch normalization           & 0.153 & 0.503 \\
			shuffles words+Glove+batch normalization & 0.150  & 0.605 \\
			full model                                   & 0.145 & 0.741 \\ \hline
		\end{tabular}
	\end{table*}
	\begin{table*}[]\scriptsize
		
		\centering    \caption{Experimental result for diversity on descriptive style. wps is the word per sentence}
		\label{table5}\footnotesize
		\begin{tabular}{p{0.5cm}|p{0.5cm}p{0.5cm}p{0.5cm}p{0.5cm}p{0.5cm}p{0.5cm}p{0.5cm}p{0.5cm}p{0.5cm}p{0.5cm}p{0.5cm}}
			\hline
			\rotatebox[origin=c]{270}{SVD factor} & \rotatebox[origin=c]{270}{total words count} & \rotatebox[origin=c]{270}{wps mean} & \rotatebox[origin=c]{270}{wps std} & \rotatebox[origin=c]{270}{BLEU1}          & \rotatebox[origin=c]{270}{BLEU2}          & \rotatebox[origin=c]{270}{BLEU3}          & \rotatebox[origin=c]{270}{BLEU4}          & \rotatebox[origin=c]{270}{METEOR}         & \rotatebox[origin=c]{270}{ROUGE\_L}       & \rotatebox[origin=c]{270}{CIDEr}& \rotatebox[origin=c]{270}{SPICE} \\ \hline
			$ \frac{1}{3}$  & 656               & 8.35     & 1.49    & 0.672  & 0.493 & 0.348 & 0.244 & 0.222  & 0.497    & 0.808 & 0.161 \\
			\\
			$ \frac{2}{3}$ & 700               & 8.45     & 1.59    & 0.674 & 0.497 & 0.354 & 0.250 & 0.226  & 0.500    & 0.827 & 0.162 \\
			\\
			1 & 791               & 8.43     & 1.53    & 0.679 & 0.501 & 0.356 & 0.252 & 0.226  & 0.501    & 0.844 & 0.166 \\ \hline
		\end{tabular}
	\end{table*}
	\normalsize
	\normalsize
	
	For evaluating diversity in generated sentences using different SVD factors, we counted words of generated captions. As shown in Table \ref{table5}, the number of words and the average of words count per sentence are decreasing as much as the factor decreases. Different factors for model aims at producing more diverse and novel captions which may not appear in the ground truths. Then, their similarity metric scores are generally less than the full model since fewer n-grams match with ground truths. Therefore, these metrics particularly represent a quality of pattern matching, instead of overall quality from the human perspective.
	
	In Figure~\ref{fig4}, all models are trained by the same set of vocabulary. The extracted vocabulary from generated captions for all models is not completely similar. This indicates diversity in generated captions by different models of MoRE.
	
	\section{Conclusion}\label{conclusion}
	We have proposed a multi-style, diverse image captioning model by using the unpaired stylized corpus. This model includes the following components:
	\begin{itemize} 
		\item A CNN as the feature extractor
		\item A Mixture of RNNs (MoRE) to embed features into a set of words
		\item An RNN that gets the output of MoRE and generates a sentence as the final output
	\end{itemize}
	Our model can generate human-like, appropriately stylized, visually grounded, and style-controllable captions. Besides, the captions made rich and diverse using a mixture of experts. The results on the COCO dataset, show that the performance of our proposed captioning model is better than previous works in case of accuracy, diversity, and styled captions. This improves the results of the previous works in BLEU, CIDEr, SPICE, ROUGE\_L, and CLF metrics. For the future works, one can consider the different ensemble methods instead of MoRE. Also to avoid overfitting, the different methods can be compared \cite{bejaniReview} to obtain the most effective one. 
\bibliography{ImageCaptioning}

\appendix

\end{document}